\documentclass[letterpaper]{article} 
\usepackage{aaai23}  
\usepackage{times}  
\usepackage{helvet}  
\usepackage{courier}  
\usepackage[hyphens]{url}  
\usepackage{graphicx} 
\usepackage{natbib}  
\usepackage{caption} 
\usepackage{algorithm}
\usepackage{algorithmic}

\usepackage{newfloat}
\usepackage{listings}
\DeclareCaptionStyle{ruled}{labelfont=normalfont,labelsep=colon,strut=off} 
\floatstyle{ruled}
\newfloat{listing}{tb}{lst}{}
\floatname{listing}{Listing}
\usepackage[table]{xcolor}

\usepackage{booktabs}

\usepackage{bm}
\usepackage{makecell}
\usepackage{wrapfig}
\usepackage{algorithm}
\usepackage{algorithmic}
\usepackage{bbm}
\usepackage{epstopdf}
\usepackage{tablefootnote}
\usepackage[normalem]{ulem}
\usepackage{color}
\usepackage{tikz}
\usepackage{bbding}

\usepackage{capt-of}

\usepackage{graphicx}
\usepackage{sidecap}

\usepackage{pifont}
\usepackage{booktabs,multirow,adjustbox}                 
\usepackage{array}
\usepackage{capt-of}
\usepackage{adjustbox}

\usepackage{bm}
\usepackage{amsmath}

\usepackage[pagebackref=false,breaklinks=true,colorlinks=true,citecolor=blue,bookmarks=false]{hyperref}
\usepackage{cleveref}  
\newcolumntype{x}[1]{>{\centering\arraybackslash}p{#1pt}}
\newcolumntype{y}[1]{>{\raggedright\arraybackslash}p{#1pt}}
\newcolumntype{z}[1]{>{\raggedleft\arraybackslash}p{#1pt}}
\newlength\savewidth\newcommand\shline{\noalign{\global\savewidth\arrayrulewidth
  \global\arrayrulewidth 1pt}\hline\noalign{\global\arrayrulewidth\savewidth}}

\title{Improving Out-of-Distribution Robustness of Classifiers\\via Generative Interpolation}
\author {
    Haoyue Bai \textsuperscript{\rm 1}\thanks{This work was done during an internship at CUHK.},
    Ceyuan Yang \textsuperscript{\rm 2},
    Yinghao Xu \textsuperscript{\rm 2},
    S.-H. Gary Chan \textsuperscript{\rm 3},
    Bolei Zhou \textsuperscript{\rm 4}
}
\affiliations {
    \textsuperscript{\rm 1} University of Wisconsin, Madison, 
    \textsuperscript{\rm 2} The Chinese University of Hong Kong, \\ 
    \textsuperscript{\rm 3} The Hong Kong University of Science and Technology,
    \textsuperscript{\rm 4} University of California, Los Angeles \\
}

\usepackage{bibentry}

\begin{document}

\maketitle

\begin{abstract}
Deep neural networks achieve superior performance for learning from independent and identically distributed (i.i.d.) data. However, their performance deteriorates significantly when handling out-of-distribution (OoD) data, where the training and test are drawn from different distributions. In this paper, we explore utilizing the generative models as a data augmentation source for improving out-of-distribution robustness of neural classifiers. Specifically, we develop a simple yet effective method called Generative Interpolation to fuse generative models trained from multiple domains for synthesizing diverse OoD samples. Training a generative model directly on the source domains tends to suffer from mode collapse and sometimes amplifies the data bias. Instead, we first train a StyleGAN model on one source domain and then fine-tune it on the other domains, resulting in many correlated generators where their model parameters have the same initialization thus are aligned. We then linearly interpolate the model parameters of the generators to spawn new sets of generators. Such interpolated generators are used as an extra data augmentation source to train the classifiers. The interpolation coefficients can flexibly control the augmentation direction and strength. In addition, a style-mixing mechanism is applied to further improve the diversity of the generated OoD samples. Our experiments show that the proposed method explicitly increases the diversity of training domains and achieves consistent improvements over baselines across datasets and multiple different distribution shifts. Code is publicly available at \url{https://github.com/HaoyueBaiZJU/generative-interpolation}.
\end{abstract}


\begin{figure}[t]
\centering
\includegraphics[width=0.48\textwidth]{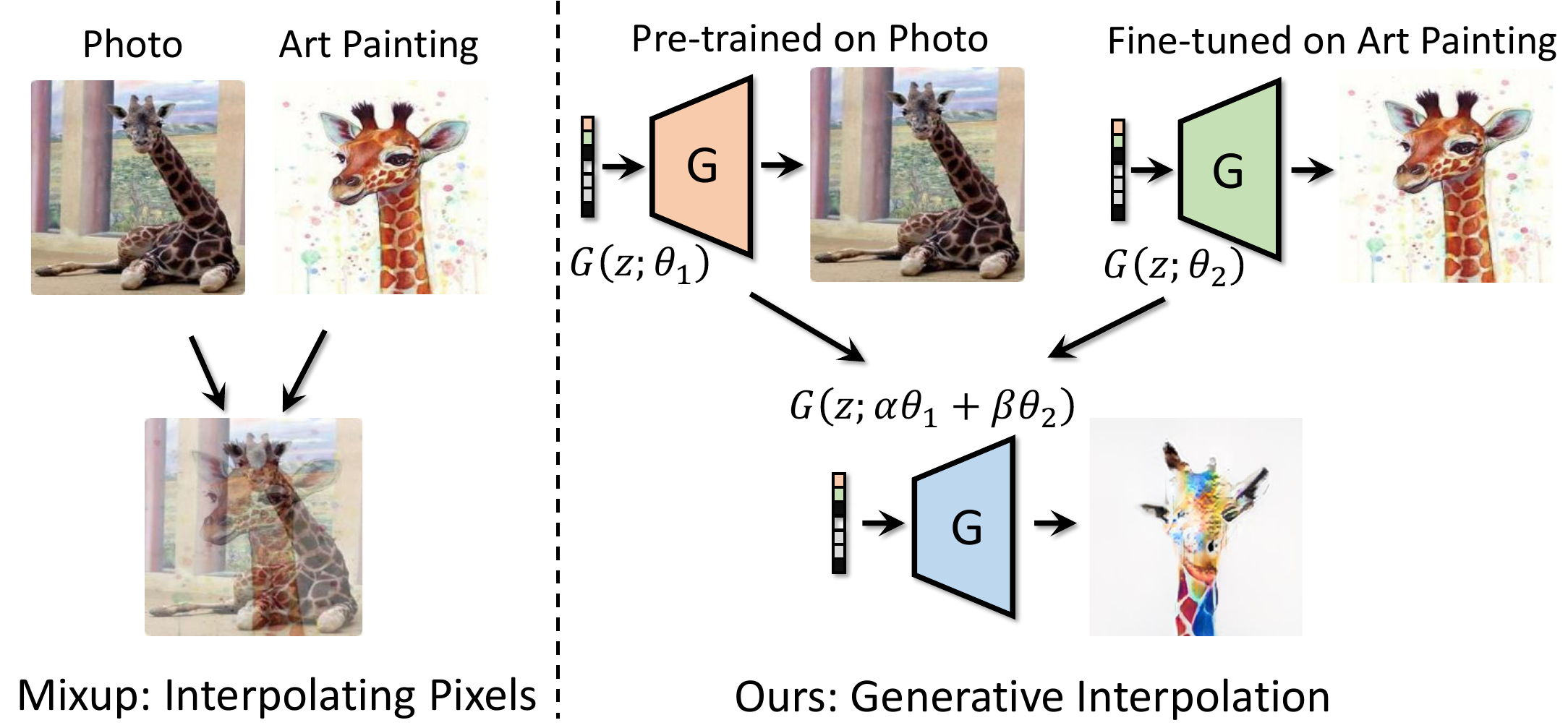}
\captionof{figure}{
Compared to the Mixup~\cite{zhang2017mixup} which interpolates the pixels of two images, the proposed Generative Interpolation method interpolates the correlated generative models trained from different domains for diverse OoD data generation. The images generated from 
interpolating model parameters can be more visually coherent. 
}
\label{fig:teaser}
\end{figure}

\section{Introduction}

Deep neural networks achieve superior performance in various applications, from computer vision~\cite{krizhevsky2012imagenet}, natural language processing~\cite{devlin2018bert}, to recommendation systems~\cite{zhang2019deep} and autonomous driving~\cite{caesar2020nuscenes}. The standard setting of model training assumes that the training and test data are drawn
independently and identically distributed (i.i.d.) from the same distribution. 
However, in the real world, the mismatch of training and test data distributions is widely observed~\cite{koh2021wilds,gulrajani2020search}, and it hurts the performance of many deep learning systems~\cite{geirhos2020shortcut}. This challenge is known as an out-of-distribution (OoD) generalization problem. Improving the robustness of classifiers against distribution shifts is still a challenging problem, as the true underlying data distributions are significantly underrepresented or misrepresented by the limited training data with selection bias~\cite{beery2018recognition}.

To improve the OoD generalization ability of classifiers, extensive efforts have been made recently. Risk regularization methods~\cite{arjovsky2019invariant,krueger2021out,ahuja2020invariant} aim to learn the invariant representation of data from different training environments by adding an invariant risk regularization.
Meta-learning for domain generalization~\cite{balaji2018metareg,dou2019domain} has been an active research area, which learns models from multiple domains that they can generalize well to unseen domains using meta-regularization.
These works, however, seldom consider improving OoD robustness in classifiers from a data augmentation perspective, which is the most straightforward and intuitive way to improve generalization ability.
Some attempts including Mixup~\cite{zhang2017mixup}, CutMix~\cite{yun2019cutmix}
have been proposed to augment the training data with 
additional synthetic examples, resulting in better generalization ability. However, mixing data in the raw pixels leads to locally ambiguous and unnatural images.
On the other hand, 
generative models such as GANs can synthesize photo-realistic images \cite{goodfellow2014generative}.
One intuitive idea is to use generative models as a data source to increase the diversity of the training data.
Previous work~\cite{jahanian2021generative,antoniou2017data} has explored this idea of generating multiple views of the same content for better representation learning.
Such an idea should be also applicable to improve the OoD robustness of classifiers, while training GAN models preserves and even amplify the correlations in the training set~\cite{tan2020improving}. Directly training classifiers on the generated data from GANs may suffer from over-fitting when the test data come from another distribution, as spurious correlations from the biased training set are unrelated to the causal reasons of target objects~\cite{arjovsky2019invariant}. 
Besides, training GAN models directly on multiple source domains may suffer from mode collapse, which means that the generator might fail to generate diverse enough images.
A proper strategy of training GAN models on multiple domains and preserving the diversity of the generated data distributions is needed for improving OoD robustness using generative models.

In this paper, we propose a simple yet effective method called Generative Interpolation to improve the OoD robustness of classifiers. The essence of the GIn method is to interpolate generative models for diverse OoD sample generation, as illustrated in Fig.\ref{fig:teaser}. The samples generated from the interpolated generative model can be much more visually coherent compared to the samples interpolated at the pixel level from Mixup~\cite{zhang2017mixup}. Our method is model-based, compared to the original Mixup which works on image pixels. 
Specifically, our approach for model interpolation is to first learn conditional generators in a pre-train and fine-tune manner from multiple source domains, and then linearly interpolate the multiple correlated networks in the parameter space such that the interpolated generator can synthesize more diverse OoD data. Both the source domain data and the generated OoD samples are used to train the robust classifiers. To be specific, we utilize StyleGAN2~\cite{karras2020training} as the data source which is pre-trained on one source domain. We then fine-tune the model on the other domains with limited distribution ranges by freezing the lower layers of the discriminator~\cite{mo2020freeze}. However, modeling the multiple source domains using correlated GANs still preserves the bias in the training set. Therefore, inspired by the network interpolation method developed for continuous style transfer~\cite{wang2019deep}, we interpolate the model parameters of the correlated generative networks trained from the multiple source domains. 
The interpolated generative models generate continuous additional vicinity of the training data with the same class, which will consistently lead to better generalization ability~\cite{simard1998transformation}. The generated OoD samples distribution also covers larger diversity ranges. Besides, the layer-wise generative representations emerge in GANs. We further perform style-mixing mechanisms to control the semantic augmentation process to alleviate the over-fitting problem to the spurious features (e.g., color) in the training set.
This helps classifiers to learn features that focus more on the shape than texture, resulting in a better OoD robustness~\cite{geirhos2018imagenet}.
The augmentation process can be controlled in fine-grained details through the interpolation coefficients. 

Our main contributions are summarized as follows:

\begin{itemize}
\itemsep0em 
    \item 
    A new method
    Generative Interpolation is developed to utilize the interpolated deep generative models for OoD generalization, where the correlated conditional generators are linearly interpolated in the parameter space to explicitly increase the diversity of source domains.
    \item We understand OoD generalization from a generative data augmentation perspective. We provide a detailed analysis of the classifiers trained on the generated OoD samples. Our practice shows that data diversity does influence OoD robustness of classifiers.
    \item Our experimental results show that our proposed framework can explicitly generate diverse OoD samples and achieves consistent improvements over baselines on the real-world OoD datasets.
\end{itemize}


\section{Related Work}

\noindent\textbf{Out-of-distribution generalization.}
OoD generalization is a fundamental problem of deep learning models, where the test data come from another distribution. OoD-Bench~\cite{ye2021ood} defines and measures the types of distribution shifts that are ubiquitous in various datasets. DomainBed~\cite{gulrajani2020search} creates a 
benchmark to facilitate reproducible domain generalization algorithms for robustness research. Multiple approaches have been proposed to improve the OoD generalization. IRM~\cite{arjovsky2019invariant} and its variants~\cite{krueger2021out,ahuja2020invariant} aims to find invariant representation from different training environments via an invariant risk regularization.
GroupDRO~\cite{sagawa2019distributionally} proposes to learn models that minimize the worst-case training loss over a set of pre-defined groups.
MLDG~\cite{li2018learning} introduces a meta-learning procedure, which simulates train and test domain shift during training.
Jigsaw~\cite{carlucci2019domain} proposes to learn the semantic labels in a supervised fashion, and jointly solve jigsaw puzzles on the same images.
In this paper, we focus on improving OoD robustness 
from a data augmentation perspective, which is the most straightforward and intuitive way to improve the generalization ability.

\noindent\textbf{Generative Adversarial Networks.}
Generative adversarial networks can synthesize photo-realistic images \cite{goodfellow2014generative}. Extensive efforts have been devoted to improving the quality of generated data~\cite{karras2017progressive,brock2018large,yang2021data}. Recent works observe layer-wise generative representations in GANs~\cite{karras2019style,karras2020training,shen2020interfacegan,xu2021generative}. Recently, some researchers take attempts to improve the fairness or acceptability of classifiers~\cite{li2021discover,mcduff2019characterizing,nguyen2017plug}.
The work of~\cite{lang2021explaining} proposes a training procedure, which incorporates the classifier model for a StyleGAN to learn a classifier-specific StyleSpace to explain a classifier.
The work~\cite{ramaswamy2021fair} introduces a GAN-based latent space de-biasing method to mitigate bias from data correlations for fair attribute classification.
In this work, we propose to 
explore interpolated generative models as a data source to 
increase the diversity of training data. However, GAN overfits easily and suffers from mode collapse training on discrete multiple source domains. Thus, it is highly non-trivial to
extend existing GANs to improve OoD robustness.

\noindent\textbf{Robustness from data augmentation perspective.} Data augmentation mechanisms augment the training data with similar but different additional virtual examples lead to better generalization ability~\cite{simard1998transformation}.
Mixup~\cite{zhang2017mixup} presents a learning principle to
generate virtual examples from a generic vicinal distribution, which trains a neural network on convex combinations of pairs of examples and labels. It has thereafter inspired some other advanced algorithms, such as Manifold Mixup~\cite{verma2019manifold}, CutMix~\cite{yun2019cutmix}, and InterpCNN~\cite{mao2019interpolated}.
DNI~\cite{wang2019deep} applies linear interpolation in the parameter space of two or more correlated networks to achieve a smooth control of imagery effects for style transfer.
The work~\cite{chai2021ensembling} uses the different views with real-world variations generated by generative models to benefit image classification.
The work~\cite{jahanian2021generative} presents that the multiview data generated by generative models can naturally be used to identify positive pairs for contrastive methods.
L2A-OT~\cite{zhou2020learning} utilizes a data generator to synthesize pseudo-novel domains data to augment 
source domains.
However, they do not consider the OoD robustness of classifiers through interpolated generative models with layer-wise generative representations.

\section{Method}

In this section, we 
present preliminaries on data augmentation and the layer-wise generative representations in GANs. 
We then introduce the detail of 
Generative Interpolation. 

\subsection{Preliminaries}

\noindent\textbf{Data Augmentation via Mixup.} 
In OoD setting, we are interested in augmenting the training data with similar but different additional samples, which can be described by the Vicinal Risk Minimization (VRM) principle~\cite{chapelle2001vicinal}. The additional samples that are drawn from the vicinity of the training data with the same class consistently result in better generalizability~\cite{simard1998transformation}. Mixup~\cite{zhang2017mixup} proposes generating virtual vectors from a generic vicinal distribution:
$\tilde{x} = \lambda x_i + (1 - \lambda) x_j$,
$\tilde{y} = \lambda y_i + (1 - \lambda) y_j$,
where $x_i$, $x_j$ are input vectors, $y_i$, $y_j$ are one-hot label encodings. The weights $\lambda$ are sampled from the Beta distribution. The neural network trained with linear interpolation of examples and corresponding labels pairs is more stable for model predictions and improves the generalization of classifiers.
However, Mixup produces locally ambiguous and unnatural samples, which misleads the model, especially for recognition~\cite{yun2019cutmix}

\noindent\textbf{Layer-wise Generative Representations in GANs.}
In order to improve the training stability and synthesis quality, the recent state-of-the-art GAN models (\emph{e.g.}, StyleGAN) introduce the layer-wise stochasticity (referred to as layer-wise generative representations in~\cite{xu2021generative}). Previous work~\cite{yang2021semantic} pointed out that such layer-wise representations could encode multi-level semantics from coarse to fine. Namely, manipulating the latent codes of some layers could merely change the corresponding semantics (\emph{e.g.}, changing the painting style while maintaining the object shape, see Fig. \ref{fig:semantic}). Besides, StyleGAN also supports the style-mixing function which swaps some layers of two latent codes such that the corresponding semantics could be re-organized. More importantly, such a style-mixing technique enables swapping the specific semantics to alleviate the correlation shifting. 
In the following section, we present how GAN models with layer-wise generative representations can be used to generate controllable virtual vicinity of the training data for better generalization ability.

\subsection{From Mixup to Generative Interpolation}

\begin{algorithm}[t]\small
\caption{Generative Interpolation for OoD generalization}
\label{alg:method}
\begin{algorithmic}[1]
\REQUIRE Training set $\mathcal{D}$, batch size $n$, learning rate $\mu$, conditional generators $\text{G}(.;\boldsymbol{\theta})$. 
\ENSURE $\boldsymbol{\theta}_1, \boldsymbol{\theta}_2,..., \boldsymbol{\theta}_K, \boldsymbol{\omega}$.
\STATE Initialize $\boldsymbol{\theta}_1, \boldsymbol{\theta}_2,..., \boldsymbol{\theta}_K, \boldsymbol{\omega}$; 
\STATE Training $\boldsymbol{\theta}_i$ on one source domain $i$; 
\STATE Obtain correlated $\boldsymbol{\theta}_1, \boldsymbol{\theta}_2,..., \boldsymbol{\theta}_K$ by fine-tuning $\boldsymbol{\theta}_i$ on other domains;
\STATE Calculate $\boldsymbol{\theta}_{\text{interp}} = \alpha_1\boldsymbol{\theta_1} + \alpha_2\boldsymbol{\theta_2} + ... + \alpha_K\boldsymbol{\theta_K}$, according to Eq. \eqref{eq:interp};
\REPEAT
\STATE Sample a mini-batch of training images $\{(x_i, y_i)\}_{i=1}^{n}$;
\STATE Sample a mini-batch of synthesized OoD samples: $x_i^{\text{syn}} \leftarrow \text{G} (x_i; \boldsymbol{\theta}_{\text{interp}})$;
\STATE $\boldsymbol{\omega} \leftarrow \boldsymbol{\omega} - \mu \cdot \nabla_{\boldsymbol{\omega}} \ell_{\text{classifier}}(\boldsymbol{\omega}, x_i, x_i^{\text{syn}})$; 
\UNTIL convergence;
\end{algorithmic}
\end{algorithm}

\begin{figure*}[ht]
    \centering
    \includegraphics[width=0.75\linewidth]{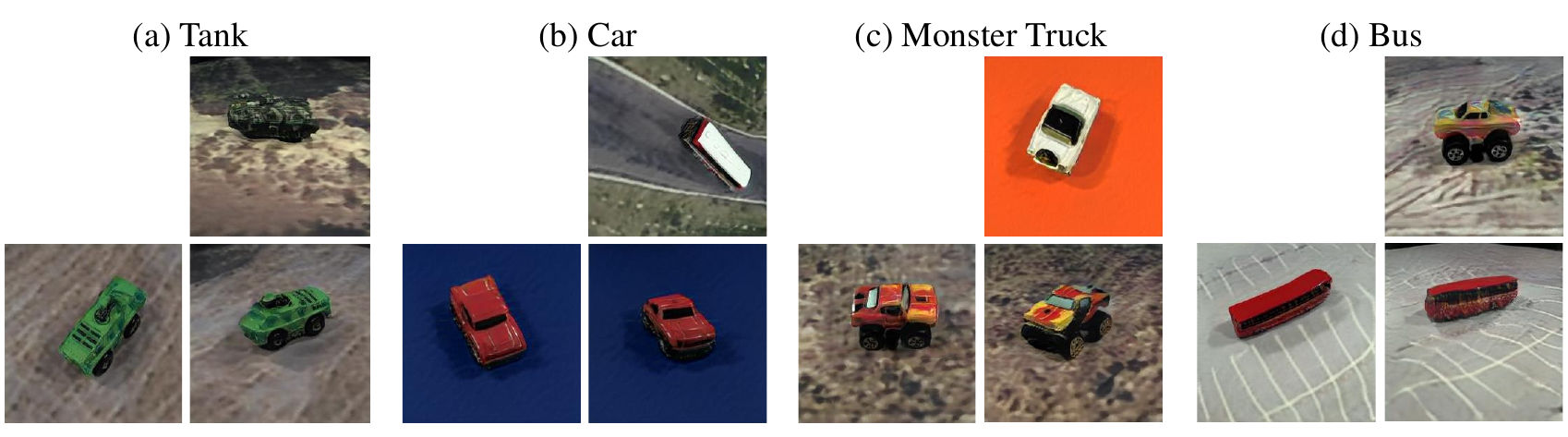}
    \caption{\textbf{Visualization of the synthesized OoD samples.} 
    Generative Interpolation manages to generate out-of-distribution samples with unseen categories and viewpoints combinations in the training set. For instance, the category and viewpoint combination for the bottom right tank in (a) is the hold-out out-of-distribution sample, which is unseen in the training set.
    However, the bottom right tank can be synthesized through our interpolated generative models.
    }
    \label{fig:OoD-sample}
\end{figure*}

Compared to the Mixup~\cite{zhang2017mixup} which interpolates the raw pixels of two images, the proposed Generative Interpolation is to generate diverse OoD samples via interpolating correlated models.
The outputs from interpolated generated models can be more coherent (see Fig. \ref{fig:teaser}).
In the OoD generalization setting, data from $K$ source domains are provided. The objective is to learn a robust classifier that can generalize well to unseen target domain. In the following descriptions, $G(\cdot;\bm{\theta_{1}}), G(\cdot;\bm{\theta_{2}}) ... G(\cdot;\bm{\theta_{K}})$ denote the $K$ correlated conditional generators for the $K$ source domains,  and $\bm{\theta_{1}}, \bm{\theta_{2}} ... \bm{\theta_{K}}$ denote the model parameters of the generators. Let $\ell_{\text{classifier}}$ be the training loss function. 

\noindent\textbf{Training GANs on Multiple Source Domains.}
To effectively train conditional GANs for improving OoD generalization in classifiers, we first train a conditional generator $G(\cdot;\bm{\theta_{i}})$ on source domain $i$. We then fine-tune the pre-trained conditional generator $G(\cdot;\bm{\theta_{i}})$ on other source domains with frozen lower layers of the discriminator to obtain the conditional generator $G(\cdot;\bm{\theta_{j}})$, where $j \ne i, 1 < j < K$. Training GANs on source domains in this pretrain and fine-tune manner is different from directly training one conditional generator on the multiple discrete source domains, which is fallible and easily result in mode collapses, as shown in our later experiments. This facility the GANs training on multiple discrete source domains and speeds up the training process. Directly training GANs on the source domains still preserves the correlations in training data. Thus, we introduce the following network interpolation in the parameter space to mitigate spurious features that are correlated but not causal for training a robust classifier. The strategy of pretraining and fine-tuning on the $K$ source domains also brings the correlated generative models where their model parameters are roughly aligned, facilitating the following network interpolation operation.

\noindent\textbf{Generative Interpolation}
In order to fuse the knowledge learned from multiple domains,
we propose to conduct linear interpolation at the model parameter space for the correlated generative networks trained from multiple source domains. The generators $G(\cdot;\bm{\theta_{1}}), G(\cdot;\bm{\theta_{2}}) ... G(\cdot;\bm{\theta_{K}})$ are mixed together via the interpolation in the parameter space as follows

\vspace{-10pt}
\small
\begin{align}\label{eq:interp}
\bm{\theta_{\text{interp}}} = \alpha_1\bm{\theta_1} + \alpha_2\bm{\theta_2} + ... + \alpha_K\bm{\theta_K}, 
\vspace{-10pt}
\end{align}
\normalsize
where $\alpha_1 + \alpha_2 + ... + \alpha_K = 1$, and there is a constraint for $\alpha_i$ that $\alpha_{i} \ge 0$.
This is a convex combination of the parameter vectors of $\bm{\theta_1}$, $\bm{\theta_2}$,..., $\bm{\theta_K}$. Diverse and continuous OoD samples can be synthesized from the interpolated generator by adjusting ($\alpha_1$, $\alpha_2$, ..., $\alpha_K$). The interpolation operation is applied on the layers in the parameter space including all the convolutional layers and normalization layers. The diversity of the OoD data can be controlled by these interpolation coefficients. In the later experiment section, we will show the generated samples and measure their diversity.

\begin{figure*}[!ht]
    \centering
    \includegraphics[width=0.78\linewidth]{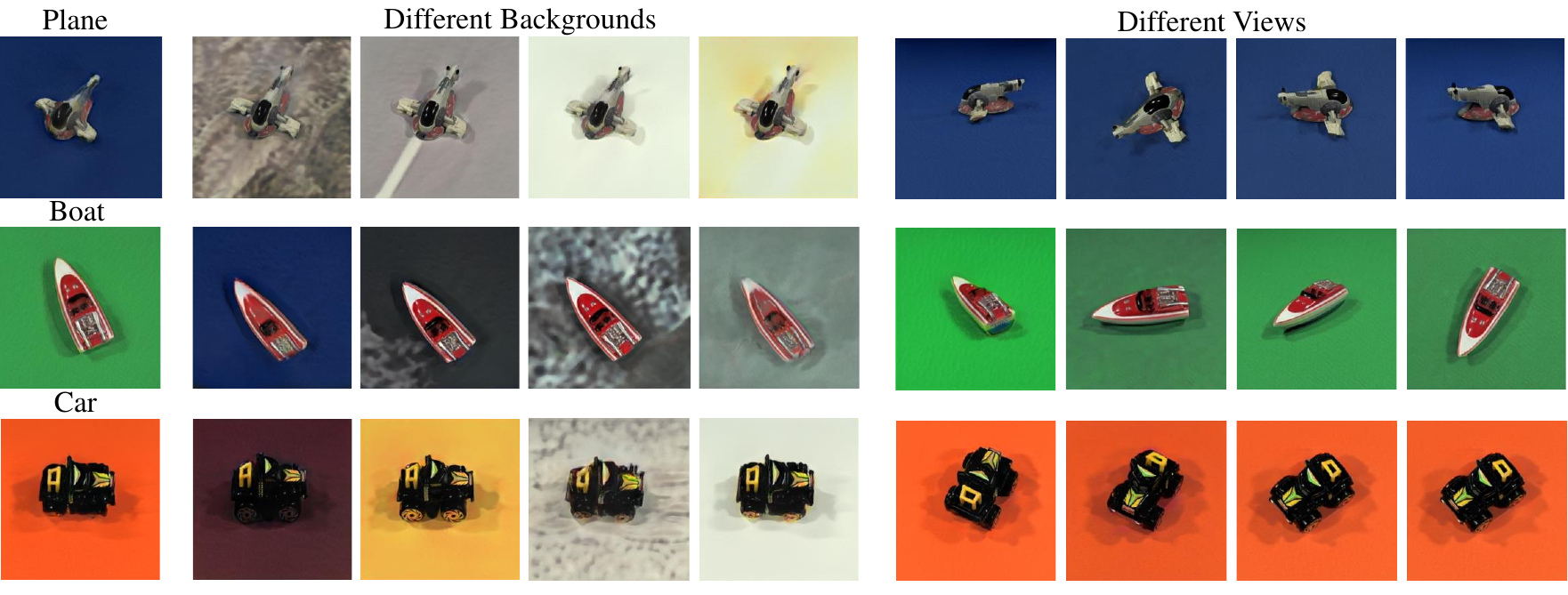}
    \caption{\textbf{Controllable synthesizing with hierarchical semantics.} Our approach Generative Interpolation manages to capture multiple semantics (\emph{e.g.}, backgrounds, views) simultaneously. Therefore, the augmentation process can be controlled in different semantic directions, resulting in a higher diversity of the generated samples. 
    }
    \label{fig:hierarchy}
\end{figure*}

\noindent\textbf{Style-mixing Strategy for Improving Data Diversity.}
After interpolating two generators, we could also leverage its layer-wise generative representation to improve the data diversity further. Specifically, given two latent codes, we could easily obtain a mixed code via the style-mixing which compose some specific layers of one with the rest layers of another. The resulting image of this mixed code could also re-compose the semantics of two syntheses correspondingly. (see Fig. \ref{fig:hierarchy}). For instance, the backgrounds of two syntheses could be exchanged by swapping the generative representations regarding the background. Therefore, the data diversity could be further improved.

\noindent\textbf{Interpolated Generator as Extra Data Source.} The interpolated generator as 
a data source can be used to increase the diversity of training set. To train the classifier and improve generalizability,
we apply the classification loss to the classifier on both real images and the synthesized OoD data:
$\min_{\omega}\, \frac{1}{N}\sum_{i=1}^{N} \mathcal{L}_{\text{classifier}}^i(\bm{\omega}, \{x_i, x_i^{\text{syn}}\})$,
\noindent where $\mathcal{L}_{\text{classifier}}$ denotes the cross-entropy loss, $\bm{\omega}$ is the parameters of the classifier, $x_i$ be the real input data, and $x_i^{\text{syn}}$ denotes the synthesized OoD samples. The stochastic gradient descent algorithm can be performed to optimize the objective. The algorithm of the proposed framework is outlined in Algorithms~\ref{alg:method}. It should be noticed that our generative model based data augmentation is diagonal to the other data augmentation methods which can be easily combined.


\begin{table}[t]
\centering
\caption{Illustration of the environment split for the iLab-2M 
\ding{55} denotes the hold-out out-of-distribution combinations.
}
\label{table:ilab2m-holdout}
\begin{tabular}{l|cccccc}
\shline
\textbf{Category}       
&\includegraphics[width=0.06\columnwidth, height=0.02\textwidth]{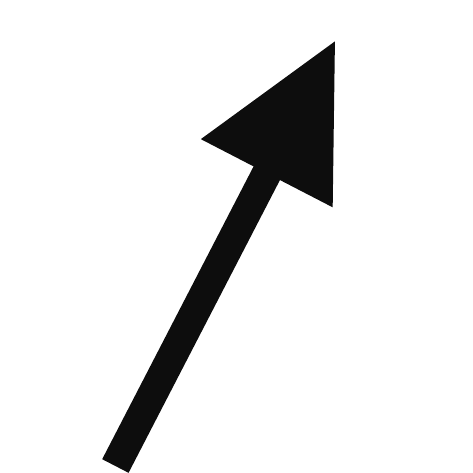}    
&\includegraphics[width=0.06\columnwidth, height=0.02\textwidth]{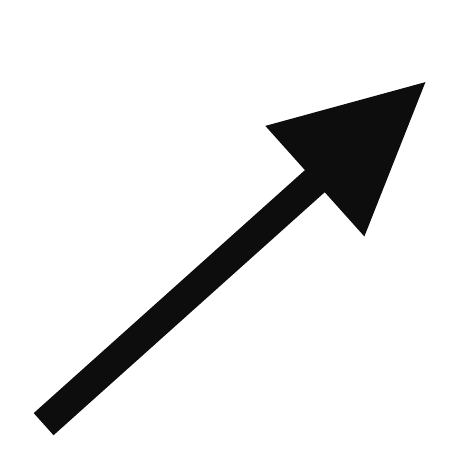}     
&\includegraphics[width=0.06\columnwidth, height=0.02\textwidth]{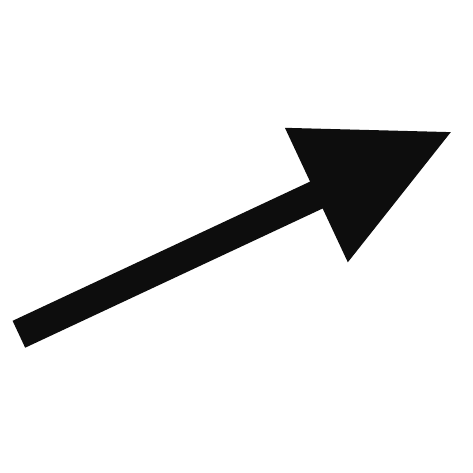}     
&\includegraphics[width=0.06\columnwidth, height=0.02\textwidth]{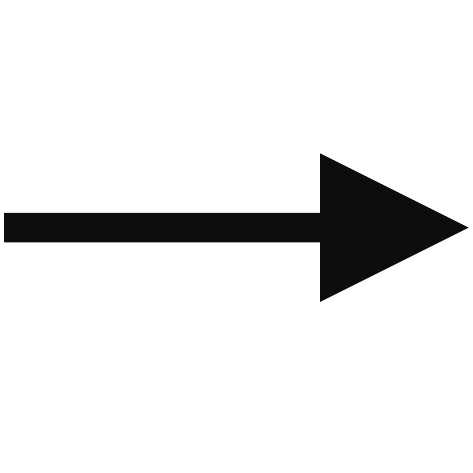}     
&\includegraphics[width=0.06\columnwidth, height=0.02\textwidth]{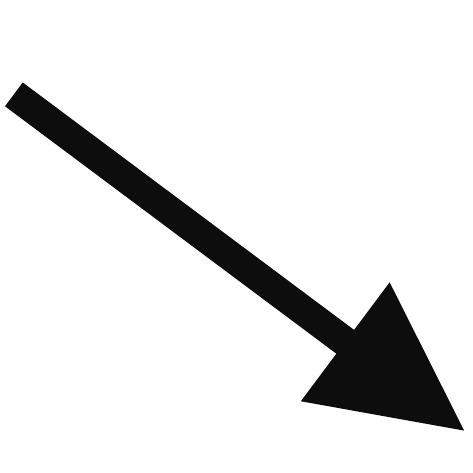}     
&\includegraphics[width=0.06\columnwidth, height=0.02\textwidth]{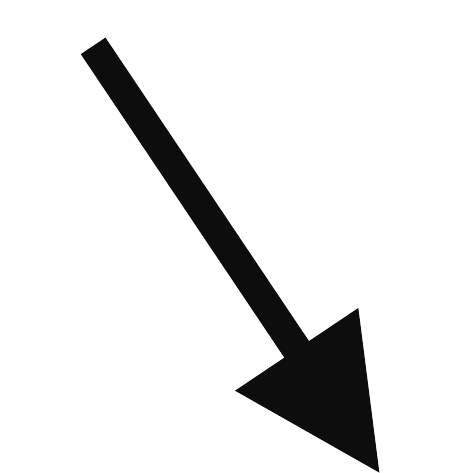}    
\\
\hline
Bus            &    &    &    &\ding{55}&    &   \\
Car            &    &    &    &    &\ding{55}&   \\
Helicopter     &    &    &    &    &    &\ding{55}   \\
Monster truck  &\ding{55}&    &    &    &    &   \\
Plane          &    &\ding{55}&    &    &    &   \\
Tank           &    &    &\ding{55}&    &    &   \\
\shline
\end{tabular}
\end{table}
\section{Experiments}

\begin{figure*}[!ht]
    \centering
    \includegraphics[width=0.78\linewidth]{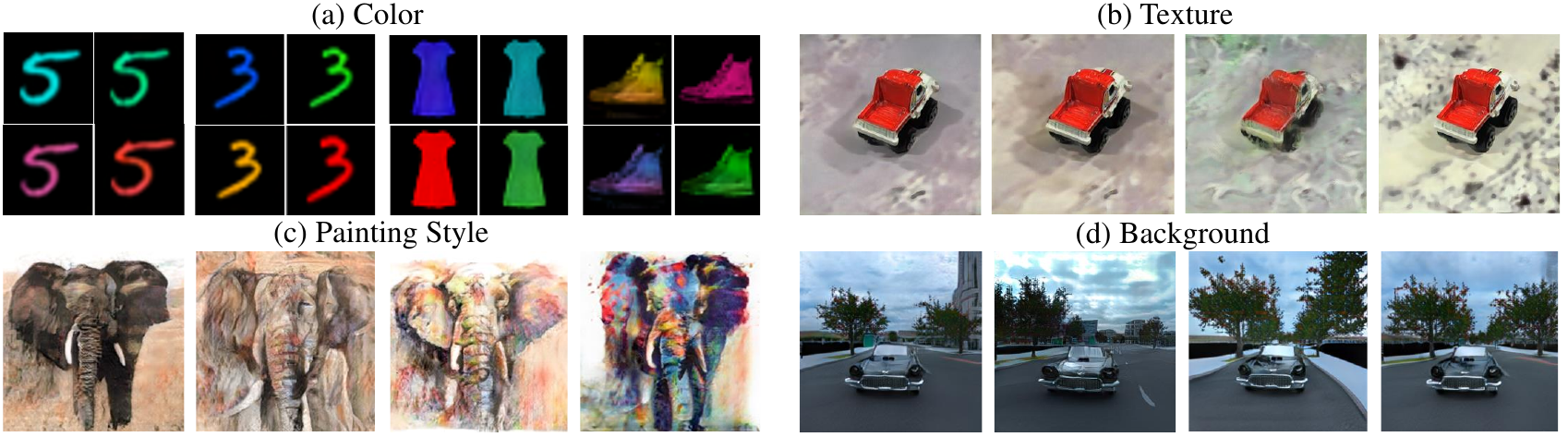}
    \caption{\textbf{Semantic identification for separate domains.}
    Generative Interpolation manages to identify the variant semantic feature (\emph{e.g.}, color, texture, painting style, and background) between separate domains. Synthesizing the samples via variant semantic direction helps the classifier for preserving the invariant features between separate domains, such as shape information.
    }
    \label{fig:semantic}
    \vspace{-1.em}
\end{figure*}

\noindent\textbf{Implementation details.} 
Generative Interpolation can be implemented on any GAN framework. In our experiment, we use the StyleGAN2~\cite{karras2020training} model to demonstrate the effectiveness of our method. 
For training the classifier on the Colored MNIST, the backbone of baseline methods is a three-layer MLP. The number of training epochs is 500, the batch size is the whole training data. The optimizer is SGD with a learning rate of 0.01. The model trained was tested at the final epoch.
The backbone network on the PACS dataset is ResNet-18. We follow the same training, validation, and test split as in the work~\cite{carlucci2019domain}.
The total training epoch is 100. The batch size is 64. 
Our framework was implemented with PyTorch 1.9.0 and CUDA 10.2. We conducted experiments on NVIDIA TITAN Xp. 
For more implementation details, see Appendix.

\noindent\textbf{Datasets.} 
We evaluate Generative Interpolation on three challenging OoD datasets with multiple different distribution shifts: Full Colored MNIST~\cite{arjovsky2019invariant}, PACS~\cite{li2017deeper} and iLab-2M datasets~\cite{borji2016ilab}.
For the Full Colored MNIST,
10 digits were colored with 10 colors based on different correlations with the labels to construct different environments.
The PACS dataset 
contains 7 categories 
and 4 domains.
We follow the same leave-one-domain-out validation protocol~\cite{li2017deeper}. 
iLab-2M is a large-scale, natural images datasets with variations in viewpoints and multiple categories. 
More details about dataset statistics 
can be found in the Appendix.

\noindent\textbf{Evaluation metric.} For evaluating OoD generalization ability, the metric is the top-1 classification accuracy. We use the metric Fr$\acute{\text{e}}$chet Inception Distance (FID)
~\cite{heusel2017gans} to evaluate the visual quality of generated data, which calculates the FID between 50,000 fake images and all the training images. Following the same setting in the work~\cite{heusel2017gans}, we use an official pre-trained Inception network to compute the FID.

\subsection{Analysis of Generative Interpolation}\label{exp:analysis}

In this section, we analyze the proposed Generative Interpolation on multiple OoD datasets including iLab-2M, PACS, and Full Colored MNIST. For the iLab-2M, the hold-out OoD combinations are shown in Tab.\ref{table:ilab2m-holdout}.

\begin{table*}[!t]
\centering
\rowcolors{1}{blue!0}{blue!4}
\captionof{table}{Performance of Generative Interpolation and OoD generalization algorithms on datasets with different distribution shifts.
The backbone for 
Full Colored MNIST dataset is MLP. The backbone for the PACS and iLab-2m dataset is ResNet-18.}
\label{table:results}
\setlength{\tabcolsep}{10pt}
\begin{tabular}{l|cccc}
\shline
\textbf{Model}    &\textbf{Colored-MNIST}  &\textbf{PACS}  &\textbf{iLab-2m}   &\textbf{Average} \\ 
\hline
IRM   &55.93     &75.92   &73.58   & 68.48\\
REx   &56.41     &77.80   &72.88   & 69.03\\
Mixup &46.75     &80.09   &70.12   & 65.65\\
ERM   &55.93     &79.05   &71.35   & 68.78\\
\hline
\textbf{Ours} &\textbf{68.04} &\textbf{83.24} &\textbf{75.30} &\textbf{75.53} \\
\shline
\end{tabular}
\end{table*}

\begin{table}[!ht]
\setlength{\tabcolsep}{2.0pt}
\centering
\rowcolors{1}{blue!0}{blue!4}
\captionof{table}{Classification accuracy on the Full Colored MNIST. 
$\textbf{Syn.}$ denotes the synthesized samples through Generative Interpolation for training the classifiers.
}
\label{table:cmnist}
\begin{tabular}{lc|cccc}
\shline
\textbf{Model}  &\textbf{Real Data}  &\textbf{Syn. 1$K$} &\textbf{Syn. 10$K$}  &\textbf{Syn. 25$K$} \\
\hline
IRM &60.04 
&60.87 
&65.33  
&\textbf{68.04 
} \\
REx &56.41 
&57.13 
&60.98 
&\textbf{63.87 
} \\
Mixup &46.75 
&49.06 
&55.50  
&\textbf{60.28 
} \\
ERM &55.93 
&56.73 
&61.00 
&\textbf{64.26 
} \\
\shline
\end{tabular}
\vspace{-1.em}
\end{table}

\noindent\textbf{Visualization of the synthesized OoD samples.} 
We visualize the generated OoD samples in Fig. \ref{fig:OoD-sample}. We find that our proposed method can synthesize out-of-distribution data, which the category and viewpoint combinations do not exist in the source domains. As shown in Tab.\ref{table:ilab2m-holdout}, certain combinations of category and view are selected as hold-out out-of-distribution samples. We can see that the category and viewpoint combination for the bottom right car in Fig. \ref{fig:OoD-sample} (b) is not included in the training set. However, Generative Interpolation manages to generate the bottom right car with unseen combinations in the training set. Therefore, training the classifier on the generated OoD samples improves the classification accuracy on the hold-out out-of-distribution test set.
In addition, several representative factors (such as backgrounds, views) exist in our proposed method, as shown in Fig. \ref{fig:hierarchy}. Several representative attributes for a single object can be identified and can be controlled to generate samples with semantic transformation. 
The generating process can be controlled through multiple semantic directions, leading to a larger diversity of the generated samples.

\noindent\textbf{Semantic identification for separate domains.} 
We have mentioned that augmenting the training data with similar but different additional virtual examples improves the generalization. The motivation is inspired by the VRM~\cite{chapelle2001vicinal} principle. 
As shown in Fig.\ref{fig:teaser}, 
Generative Interpolation is indeed able to generate more realistic images than the Mixup
algorithm with preserved class labels. 
We observe that some attributes, e.g. colors, styles,
are successfully being changed with shape remains the same, see Fig.\ref{fig:semantic}. This may be due to the emergence of semantic hierarchy in the latent space of interpolated generative models. The classifiers trained with these mixed samples help to alleviate the over-fitting problem to the spurious feature (\emph{e.g.}, colors) 
and focus more on the shape feature of digits. The neural network models, which focus more on shape than texture, have better generalizability~\cite{geirhos2018imagenet}.

\begin{figure}[t]
    \centering
    \includegraphics[width=0.75\linewidth]{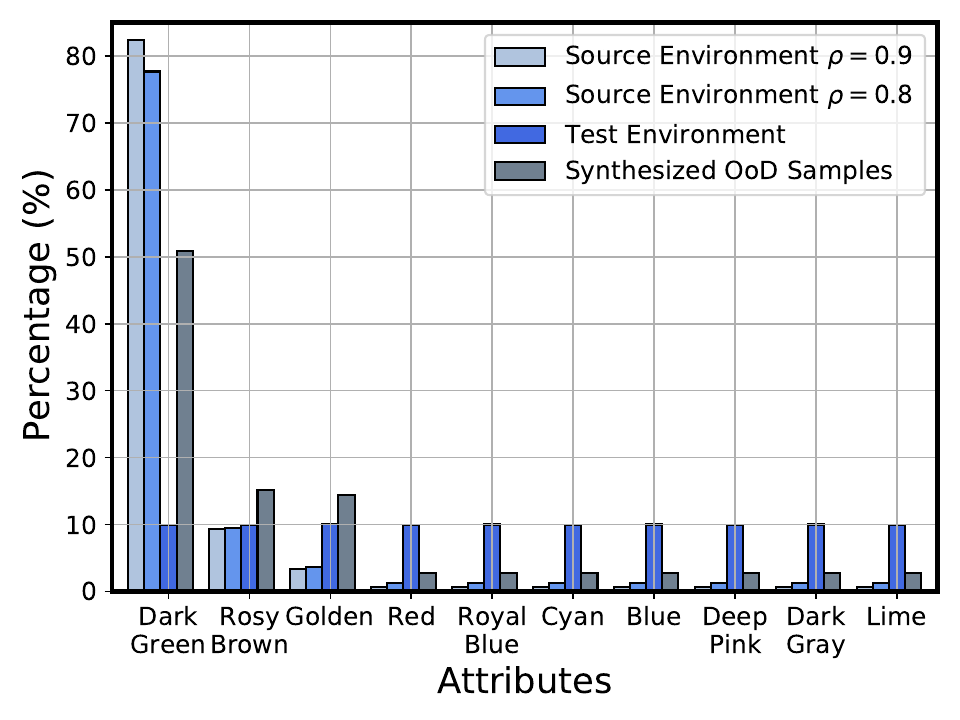}
    \caption{\textbf{Quantifying the generated OoD samples.}
    \textbf{Statistical analysis of the synthesized OoD samples.} The data distribution percentage over different attributes for source environment $\rho=0.9$, source environment $\rho=0.8$, unseen domain, and synthesized OoD samples. 
    }
    \label{fig:quan}
    \vspace{-1.em}
\end{figure}

\noindent\textbf{Statistical analysis of the synthesized OoD samples.} 
To check whether the distribution of the synthesized samples is different from the training set, we plot the different color type's percentages for the training environments, test environment, and the synthesized OoD samples for the Full Colored MNIST, in Fig. \ref{fig:quan}. We observe that the distribution over the different color types of synthesized OoD samples tends to approach the test environment, compared with the distribution of the training environment.
Furthermore, we use the Kullback-Leibler (KL) divergence, which is a general distance measure for distributions, to quantify the degree of distribution shifts among training set, test set, and synthesized data. The distance between the synthesized samples and test environment (1.01) is much smaller than the distance between the synthesized samples and source environment $\rho=0.9$ (2.89), which supports that our proposed method can generate out-of-distribution samples.

\noindent\textbf{Visual quality of the synthesized samples.} 
The quantitative results of the visual quality in terms of FID for the synthesized data are shown in Tab.\ref{table:fid} for the PACS dataset.
We observe that training GAN models directly on multiple source domains are not easy, 
while the synthesizing quality is substantially improved via the proposed pretrain and fine-tuning strategy. Specifically, the visual quality 
for the sketch domain achieves 20.2 by pretraining on the photo domain and fine-tuning to the sketch domain.
It should be emphasized that indeed the synthesized images are not as photo-realistic as the real images. However, our purpose here is not to achieve high synthesis quality, instead, we want the synthesized images to carry enough discriminative information for training classifiers. As shown in the following OoD generalization experiments, the high-level features underlying the synthesized images are \textit{good enough} for a neural network classifier to recognize different categories thus improving the OoD generalization of the trained classifier.

\subsection{OoD Generalization Results and Discussion}\label{exp:results}

In this section, we evaluate the proposed Generative Interpolation and analyze its results on three datasets: Full Colored MNIST, PACS, and iLab-2M, which represent different aspects of distribution shifts in the OoD problem.

The results for the PACS dataset are shown in Tab.\ref{table:results}. 
Our method achieves 
the SOTA performance when using the ResNet-18 as the backbone network no matter whether the backbone is pre-trained or randomly initialized. 
This PACS dataset considers realistic generalization scenarios with distribution shifts in styles. 
Generative Interpolation achieves $83.24\%$ average accuracy, compared with some representative OoD algorithms, such as MTL~\cite{blanchard2017domain} ($79.50\%$) and Mixup~\cite{zhang2017mixup} ($80.09\%$). 
The poor performance of Mixup might be due to the locally ambiguous and unnatural samples when performing the interpolation on the raw pixels.
The superior performance of 
Generative Interpolation also confirms the great potential of improving the classifier's OoD robustness by using interpolated generative models as an extra data source.

\begin{table*}[ht]
\begin{minipage}{0.48\linewidth}
\rowcolors{1}{blue!0}{blue!4}
\centering
\captionof{table}{Experiments on advanced data augmentations on fashion MNIST. The backbone for the baselines is MLP.}
\label{table:aug}
\begin{tabular}{l|c}
\shline
\textbf{Model}  &\textbf{Accuracy} \\ 
\hline
ERM                       & 45.06  
\\
ERM + Syn                 & 61.12  
\\
Cutout + Syn              & 62.29 
\\
CutMix + Syn              & 62.34  
\\
\shline
\end{tabular}
\end{minipage}
\hspace{1em}
\begin{minipage}{0.48\linewidth}
\rowcolors{1}{blue!0}{blue!4}
\centering
\captionof{table}{Visual quality in terms of FID (lower is better)  for the synthesized samples on PACS dataset.
}
\label{table:fid}
\setlength{\tabcolsep}{10pt}

\begin{tabular}{l|c}
\shline
\textbf{Sketch Domain}  &\textbf{FID} \\ 
\hline
Joint Photo and Sketch & 82.3 \\
Sketch                 & 96.3 \\
Cartoon to Sketch (ours) & \textbf{32.9} \\
Photo to Sketch (ours)   & \textbf{20.2} \\
\shline
\end{tabular}
\end{minipage}
\end{table*}

Tab.\ref{table:cmnist} demonstrate the advantages of our method.
The proposed method achieves consistent improvements over baselines including risk regularization methods, such as REx and IRM. Specifically, IRM~\cite{arjovsky2019invariant} combined with interpolated GAN method achieves better 
accuracy (68.04\%), even compared with the REx~\cite{krueger2021out} method. Our method further improves the performance in Colored MNIST by synthesizing extra data through interpolating generative models to augment the training set. 
Noticing that all the four baselines show better OoD accuracy when combined with the proposed approach.
These results confirm the possibility of improving OoD generalization ability through interpolated generative models.

The recently proposed iLab-2M dataset is a large-scale object-centric image dataset. Results in Tab.\ref{table:results} 
demonstrate the advantages of our methods. In our implementation, the generalization accuracy is much better than previous OoD algorithms ERM (71.35\%), IRM 
(73.58\%), REx
(72.88\%) with ResNet-18 backbone, which are the best OoD approaches before our approach. In addition, Mixup
achieved 70.12\% accuracy on iLab-2M, indicating that mixing up (interpolating) data in the raw pixel may not be able to correct the spurious correlation between irrelevant features to predict category. As shown in Fig. \ref{fig:OoD-sample}, our proposed method manages to synthesize out-of-distribution samples with unseen categories. The results 
also demonstrate that training the classifier on the synthesized samples improves classification on the hold-out out-of-distribution accuracy.

\subsection{Ablation Study and Sensitivity Analysis}\label{exp:ablation}

We conduct an ablation study on different quantities of generated data. 
We conduct experiments to investigate combining our approach with existing advanced data augmentations. 
More ablation studies are available in the Appendix.

\noindent\textbf{Experiments on the different quantities of generated data.} 
We conduct comprehensive ablation studies on the different quantities of synthesized OoD samples. The results of the training classifier are shown in Tab.\ref{table:cmnist} for the Full Colored MNIST dataset. We observe that IRM achieves $60.04\%$ out of distribution accuracy on the real data, which is much lower than training together with $25K$ synthetic data ($68.04\%$). In addition, we also find that the OoD accuracy is increased from $56.73\%$ with $1K$ generated OoD samples to $64.26\%$ with $25K$ generated data when combining with the ERM baseline. This may be because the interpolated generative models are able to generate OoD samples and increase the data diversity of the training environment.

\noindent\textbf{Experiments on combining with different baselines.}  
We change current ERM baseline to other representative OoD algorithms, such as Mixup,
IRM,
and REx
(see Tab.\ref{table:cmnist}). Our proposed approach achieves consistent improvement in terms of OoD accuracy on the four baselines. We tried REx with augmented data on Colored MNIST, the result is $63.87\%$ which is much higher than the baseline, which is $56.41\%$ out-of-distribution accuracy. This shows that training a classifier with synthesized OoD samples is essential for increasing OoD robustness. These results also demonstrate that our proposed model-based data augmentation can be easily combined with existing OoD algorithms.

\noindent\textbf{Experiments on advanced data augmentations.} 
To check whether the proposed Generative Interpolation can be used together with existing data augmentation schemes, we present the performance on combining with different advanced data augmentations. The results are shown in Tab.\ref{table:aug}. These experiments were repeated five times.
We find that the proposed Generative Interpolation approach performs well when combined with advanced data augmentations such as CutMix, Cutout. 
We can see from the table that CutMix combined with Synthetic data achieves 62.34\% out-of-distribution accuracy, Cutout combined with Synthetic data achieves 62.29\%. These results show that our generative model-based data augmentation is orthogonal to the other data augmentation methods which can be easily combined.

\noindent\textbf{Experiments on imbalanced domains.} Our proposed approach with a pretrain and fine-tune manner facilitates training the generative models on imbalanced source domains. We present the visual quality of the synthesized samples 
in Tab.\ref{table:fid}. The visual quality of the synthesized samples is 96.3 when directly trained on the sketch domain. We observe that pretraining on the photo domain and fine-tuning on the sketch domain achieves 20.2 in terms of FID. This may be because if one domain has very less semantics tends to suffer from mode collapse. Pretraining on domains that contain more diverse semantics (e.g., photo domain) learn better representations and converge faster when fine-tuning on domains with less semantics (e.g., sketch domain).

\section{Conclusion}

In this paper, we develop the method called Generative Interpolation to utilize the interpolated generative models 
as a data source to increase the diversity of the training domains for improving OoD robustness. 
The method first learns correlated GAN models on multiple source domains in a pretrain and fine-tune manner.
Then multiple correlated generators are linearly interpolated in the parameter space for generating diverse OoD samples. A style-mixing mechanism is further used to increase the diversity of the output OoD samples. Extensive experiments show that our proposed method explicitly generates diverse OoD samples and achieves consistent improvements over baselines and previous methods on several OoD generalization datasets.

\clearpage

\bibliography{aaai23}

\clearpage

\appendix

\small


\section{More Ablation Study Results} \label{appendix:ablation}

\noindent\textbf{Ablation study on different components.}
We conduct an ablation study on the network interpolation and style-mixing mechanism.
We tried directly applying StyleGAN2
for the multiple source domains on the PACS dataset. As shown in Tab.\ref{table:ablation}, the result is $47.82\%$. This may be because training conventional GAN models for data augmentation still preserves the correlations in the training data. 
This shows that the proposed framework is needed to improve OoD robustness via data augmentation. 
The average accuracy on PACS without network interpolation and style-mixing is $47.82\%$.
The average accuracy is improved after performing the network interpolation mechanism, which is $49.83\%$ on the PACS dataset. The baseline algorithm is ERM, and we use StyleGAN2
as the baseline generative model. 
This shows the effectiveness of network interpolation to facilitate the semantic augmentation process. A style-mixing mechanism is further proposed to increase the diversity of the generated data and achieve a higher average accuracy of $50.42\%$, which shows the effectiveness of the style-mixing mechanism. This may be due to the layer-wise generative representations being leveraged to generate controllable virtual vicinity of the training data.

\setlength{\tabcolsep}{5pt}
\begin{table}[h]
\rowcolors{1}{blue!0}{blue!4}
\centering
\captionof{table}{Ablation study on the components in Generative Interpolation. Out-of-distribution accuracy serves as the evaluation metric.}
\label{table:ablation}
\setlength{\tabcolsep}{4pt}
\begin{tabular}{ccc|c}
\shline
\textbf{StyleGAN2}  &\textbf{Interpolation}  &\textbf{Style-mixing} &\textbf{Accuracy}\\
\hline
            &           &           &46.82    \\
\ding{51}  &           &           &47.82    \\
\ding{51}  &\ding{51} &           &49.83    \\
\ding{51}  &\ding{51} &\ding{51} &\textbf{50.42}    \\
\shline
\end{tabular}
\end{table}

\noindent\textbf{Ablation study on data diversity.} We present more ablation studies to test whether the larger data diversity can achieve better out-of-distribution generalization performance or not. The results on the iLab-2M dataset are shown in Tab.\ref{table:iLab-2M}. The scenarios of category as labels and the rotation as labels are evaluated. 
Tab.\ref{table:iLab-2M} presents the out-of-distribution performance for fixed dataset size, but different data diversity ratios. For fixed dataset size, increasing the diversity of in-distribution combinations makes the classifier perform better for the out-of-distribution scenarios. We observe that the OoD generalization performance achieves 74.66\% with diversity ratio 1, compared with 52.19\% OoD accuracy with lower data diversity $\alpha=2/3$. This demonstrates that the data diversity does influence the out-of-distribution robustness of classifiers.

\setlength{\tabcolsep}{4.5pt}
\begin{table}[t]
\rowcolors{1}{blue!0}{blue!4}
\centering
\caption{Classification accuracy on the iLab-2m dataset with different data diversity ratios. The backbone for the baseline is ResNet-18. `\ding{51}' denotes the results of in distribution accuracy. `\ding{55}' denotes the results of out of distribution accuracy.}
\label{table:iLab-2M}
\begin{tabular}{cc|ccc}
\shline
Model & In Distribution  &$\alpha = 1/3$ &$\alpha=2/3$ &$\alpha=1$\\
\midrule
ERM & \ding{51}   
&100  &100  &100 \\
(category) &\ding{55}      
&23.18 &52.19 &74.66 \\
ERM & \ding{51}   
&100 &100 &100\\
(rotation) &\ding{55}      
&33.55 &67.60 &88.55 \\
\shline
\end{tabular}
\vspace{-1.em}
\end{table}

\begin{figure*}[ht]
\centering
\includegraphics[width=0.95\textwidth]{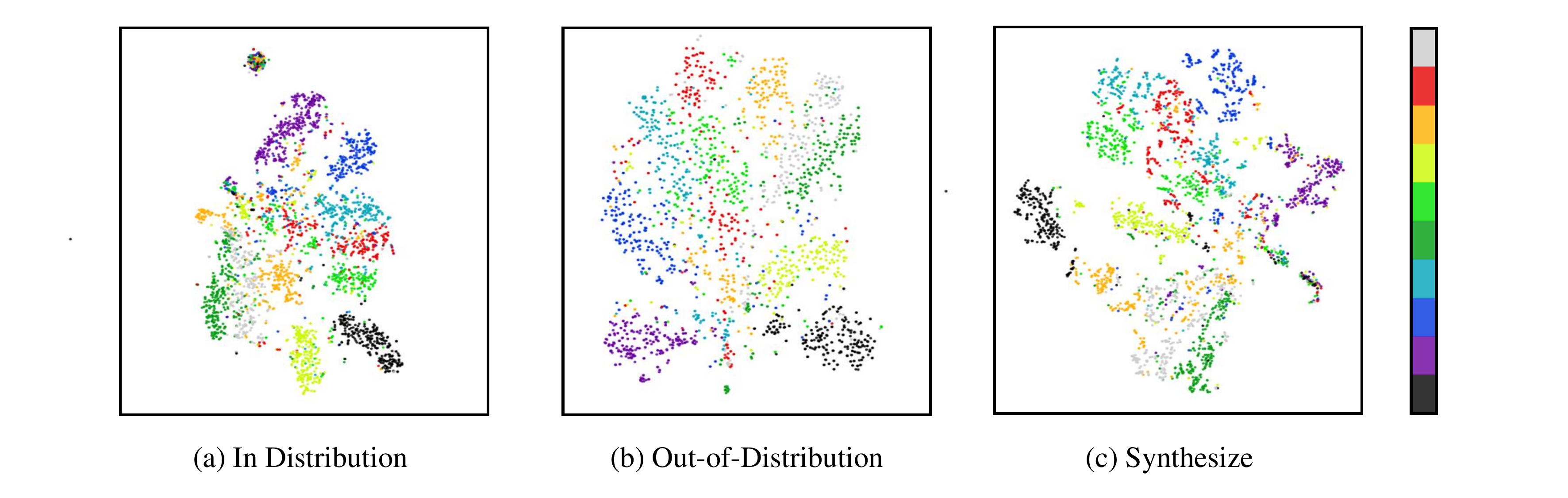}
\caption{\textbf{T-SNE visualization} for the (a) in distribution, (b) out-of-distribution, and (c) synthesized data.}
\label{fig:t-sne}
\end{figure*}

\begin{figure*}[ht]
\centering
\includegraphics[width=0.7\textwidth]{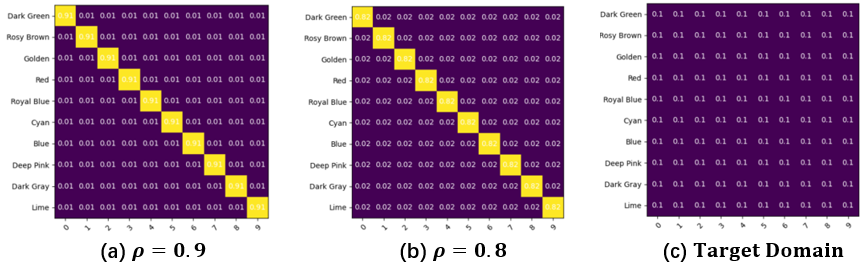}
\caption{\textbf{Illustration of the Full Colored MNIST dataset.} The digits were colored with 10 colors based on different correlations with labels to construct different environments, i.e., $80\%$ and $90\%$ for training environments and $10\%$ for the test environments.
}
\label{fig:cmnist-illustrate}
\end{figure*}

\setlength{\tabcolsep}{5pt}
\begin{table*}[ht]
\rowcolors{1}{blue!0}{blue!4}
\centering
\caption{Classification accuracy on the fashion MNIST. The backbone for the baselines is MLP. The baselines are implemented by ourselves.}
\label{table:fashion-mnist-appendix}
\begin{tabular}{lcc|cccc}
\shline
Model & In Distribution  &Real Data  &Syn. 1$K$  &Syn. 10$K$ &Syn. 20$K$\\
\midrule
ERM & \ding{51}   
&94.41$\pm$0.08 &92.55$\pm$0.09 &89.91$\pm$0.05 &88.47$\pm$0.03 \\
~ &\ding{55}      
&45.06$\pm$0.14 &45.46$\pm$0.38 &57.18$\pm$0.21 &\textbf{61.12$\pm$0.33}  \\
\shline
\end{tabular}
\vspace{-1.em}
\end{table*}

We also conduct experiments and ablation studies on more OoD datasets such as the Fashion MNIST dataset.
Our method achieves much better performance, in terms of out-of-distribution accuracy on the fashion MNIST compared with ERM baseline, see Tab.\ref{table:fashion-mnist-appendix}. Specifically, ERM combined with our proposed augmentation method achieves $61.12\%$ when synthetic data is $20,000$, which is much higher than the ERM baseline ($45.06\%$). This may be because the synthetic data facility the classifier disregards spurious features that are correlated but not causal for prediction. 
Besides, we also conduct experiments on the different quantities of generated data on this dataset. We observe that the OoD accuracy is increased from 45.46\% with $1K$ synthesized samples to 57.18\% with $10K$ synthesized samples. This might be because our proposed method manages to generate OoD samples and increase the diversity of data for training classifiers.

\section{T-SNE Visualization} \label{appendix:t-SNE}

\noindent We also use the interpretability method t-SNE to explain the distribution of the generated data, as shown in Fig.\ref{fig:t-sne}. We do t-SNE to visualize the embedding of (a) in distribution, (b) out-of-distribution, and (c) distribution of generated data. We take the Colored MNIST dataset for example. Different colors in the t-SNE visualization denote different categories. It can be seen that the patterns for the in distribution data are more contract, compared with out-of-distribution and generated data. We observe that the patterns of generated data cover larger ranges, and are closer to the out-of-distribution data. This demonstrates that our proposed method can generate OoD samples with a large diversity.

\section{Implementation Details} \label{appendix:details}
\noindent We conduct a fair comparison of our method with various OoD generalization algorithms and SOAT GAN methods on challenging OoD datasets. For our proposed method, the learning rate for the generation process is 0.0025. The optimizer for training the conditional generator is. For training the classifiers, we use the SGD optimizer with an initial learning rate of 0.01 for the Full Colored MNIST and the Colored Fashion MNIST. For the PACS dataset, the batch size is 64. The optimizer is SGD. We conduct hyper-parameter optimization for all the baseline methods and compare our proposed method with their performance under the best hyper-parameters. The results for the Colored MNIST and Colored Fashion MNIST have averaged over 5 runs with the best set of hype parameters.

For the Full Colored MNIST, we use ten colors for all the data with 10 digits: Dark Green ([0, 100, 0]), Rosy Brown ([188, 143, 143]), Golden ([255, 215, 0]), Red ([255, 0, 0]), Royal Blue ([65, 105, 225]), Cyan ([0, 225, 225]), Blue ([0, 0, 255]), Deep Pink ([255, 20, 147]), Dark Gray ([160, 160, 160]), Lime ([0, 255, 0]). The image resolution is $32 \times 32$. The dataset size is 60000 images. The correlation is different during training and test time.

\begin{figure*}[ht]
    \centering
    \includegraphics[width=1.0\linewidth]{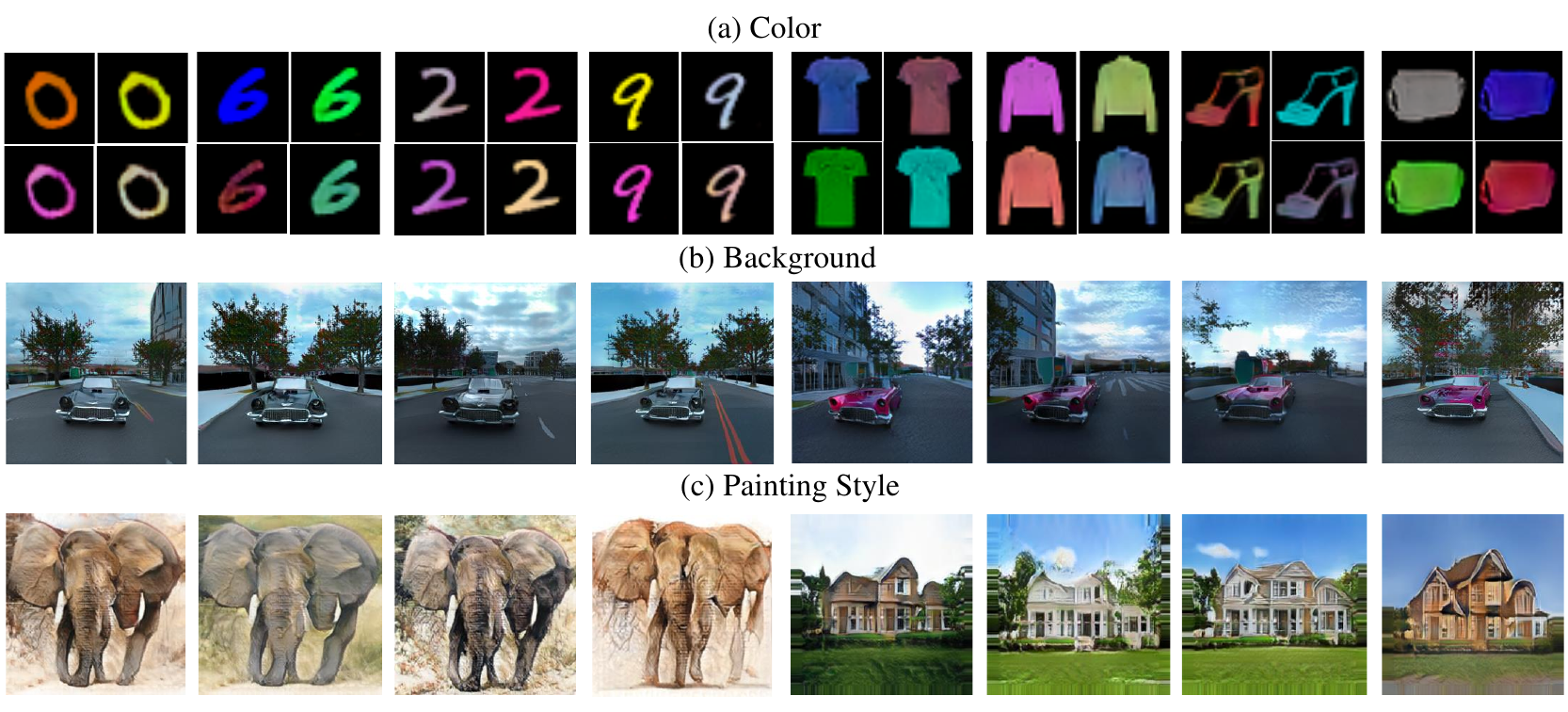}
    \caption{\textbf{Semantic interpolation.} Our proposed method can identify and transfer variant semantic features such as color, background, and painting style, which helps the classifier to preserve invariant features (e.g., shape information).
    }
    \label{fig:consist}
\end{figure*}

\begin{figure*}[ht]
    \centering
    \includegraphics[width=0.68\linewidth]{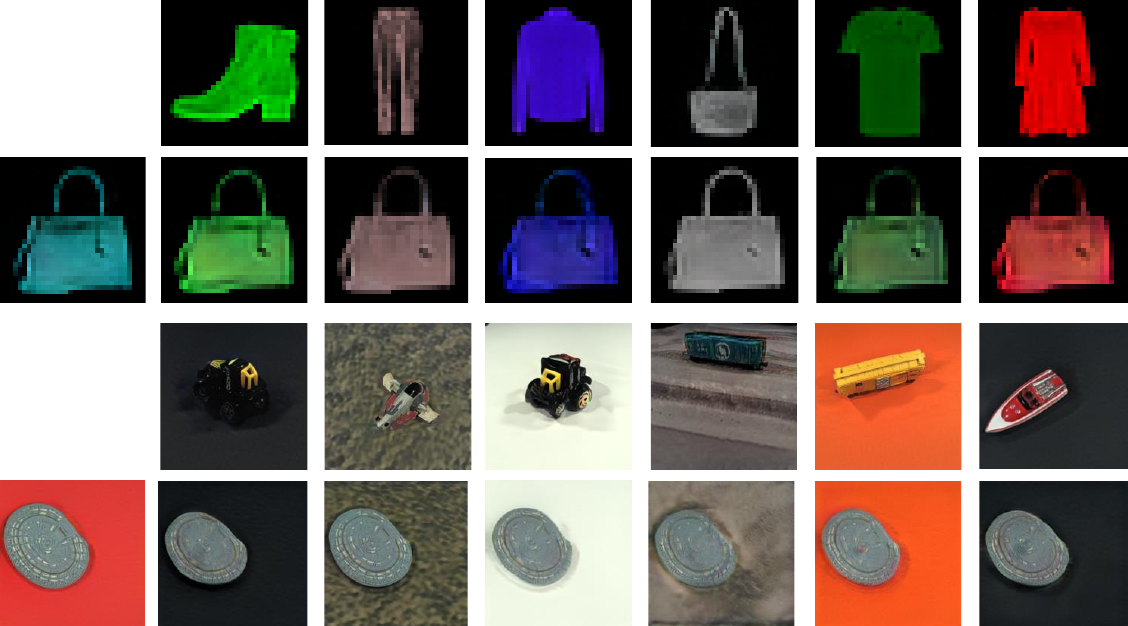}
    \caption{\textbf{Generating OoD samples with variation factor combinations.} Our proposed method is able to synthesize images with attributes only available with features from another category.
    }
    \label{fig:factor}
\end{figure*}

For the Colored Fashion MNIST with 10 colors and 10 categories (T-shirt, Trouser, Pullover, Dress, Coat, Sandal, Shirt, Sneaker, Bag, Ankle boot). This is a more challenging task than the Colored MNIST. We color the objects based on different correlations to construct different environments. Similar to the Full Colored MNIST, We set a one-to-one digit-color relationship and set the bias coefficient to $\rho=0.9$ and $\rho=0.8$.

PACS is a widely used OoD dataset. This dataset contains 9991 images with 7 categories (dog, elephant, giraffe, guitar, horse, house, person) and 4 domains (photo, art painting, cartoon, sketch).
The original images with $227 \times 227$ resolution were padding to $256 \times 256$ when training the generative models.
The implementation details of the pretraining and fine-tuning stage for the pacs dataset are: 1) Target domain photo: pretraining on the cartoon domain, fine-tuning to sketch, and art painting. 2) Target domain art painting: pretraining on the photo domain, fine-tuning to sketch, and cartoon. 3) Target domain sketch: pretraining on the photo domain, fine-tune to art painting, and cartoon. 4) Target domain cartoon: pretraining on the photo domain, fine-tune to art painting and sketch.

iLab-2M includes toy vehicle objects under variants of viewpoint, lighting, and background. This dataset contains a total of two million images: 1.2M training, 270K validation, 270K test images with six different azimuth viewpoints, and five different zenith angles (total 30).
This dataset has 15 categories originally. In our experiments, we chose a subset of six categories: Bus, Car, Helicopter, Monster Truck, Plane, Tank, and six viewpoints labels.

\begin{figure*}[t]
    \centering
    \includegraphics[width=0.95\linewidth]{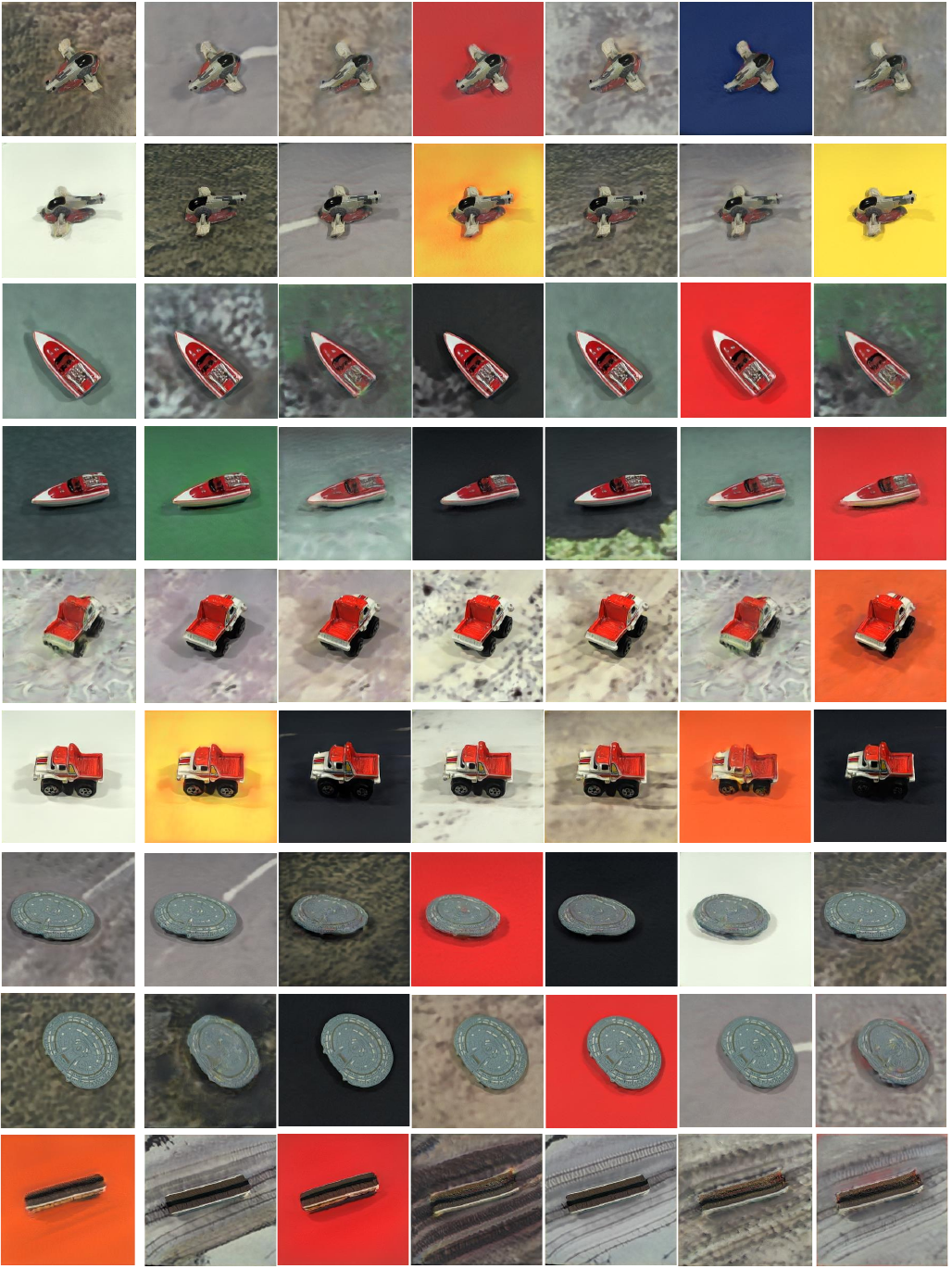}
    \caption{\textbf{Semantic interpolation.} Our approach manages to capture and transfer domain variant semantics such as the background. We provide more visualization results for the generated OoD samples with different categories (e.g., helicopter, boat, train). 
    }
    \label{fig:iLab-2M}
\end{figure*}

\section{Additional Visualization Results} \label{appendix:vis}

\noindent We provide more visualization results for the generated OoD data as shown in Fig. \ref{fig:consist}, Fig. \ref{fig:factor}, and Fig. \ref{fig:iLab-2M}.

\end{document}